\documentclass[letterpaper, 10 pt, conference]{ieeeconf}

\IEEEoverridecommandlockouts                              

\let\labelindent\relax

\usepackage{microtype}

\usepackage{enumitem}
\usepackage{svg}
\usepackage{subcaption}
\usepackage{cite}

\usepackage{mathtools,amsmath, amssymb,xfrac,nicefrac, amsfonts}
\usepackage{float}
\usepackage[ruled,vlined,linesnumbered]{algorithm2e}
\usepackage{graphicx}
\usepackage{textcomp}
\usepackage{xcolor}
\usepackage[capitalize]{cleveref}
\usepackage{tabularx}
\usepackage{booktabs}

\usepackage{siunitx}

\DeclareMathAlphabet{\pazocal}{OMS}{zplm}{m}{n}

\newcommand{\mat}[0]{\begin{bmatrix}}
\newcommand{\mate}[0]{\end{bmatrix}}

\newcommand{\va}{\mathbf{a}}

\newcommand{\vh}{\mathbf{h}}

\newcommand{\vM}{\mathbf{M}}

\newcommand{\vP}{\mathbf{P}}

\newcommand{\vw}{\mathbf{w}}
\newcommand{\vx}{\mathbf{x}}

\newcommand{\vphi}{\boldsymbol{\phi}}

\newcommand{\cC}{\mathcal{C}}
\newcommand{\cD}{\mathcal{D}}
\newcommand{\cE}{\mathcal{E}}
\newcommand{\cF}{\mathcal{F}}
\newcommand{\cG}{\mathcal{G}}

\newcommand{\cI}{\mathcal{I}}

\newcommand{\cM}{\mathcal{M}}

\newcommand{\cO}{\mathcal{O}}
\newcommand{\cP}{\mathcal{P}}

\newcommand{\cS}{\mathcal{S}}

\newcommand{\cV}{\mathcal{V}}
\newcommand{\cW}{\mathcal{W}}

\DeclareMathOperator*{\argmin}{arg\,min}            %

\makeatletter
\newcommand*\wt[1]{\mathpalette\wthelper{#1}}
\newcommand*\wthelper[2]{%
        \hbox{\dimen@\accentfontxheight#1%
                \accentfontxheight#11.3\dimen@
                $\m@th#1\widetilde{#2}$%
                \accentfontxheight#1\dimen@
        }%
}

\newcommand*\accentfontxheight[1]{%
        \fontdimen5\ifx#1\displaystyle
                \textfont
        \else\ifx#1\textstyle
                \textfont
        \else\ifx#1\scriptstyle
                \scriptfont
        \else
                \scriptscriptfont
        \fi\fi\fi3
}
\makeatother

\SetKw{Continue}{continue}
\SetKw{Break}{break}

\title{
 
\LARGE \bf Learning Semantic Priorities for Autonomous Target Search
}
\author{Max Lodel, Nils Wilde, Robert Babu\v{s}ka, Javier Alonso-Mora
\thanks{This work was supported by the National Police of the Netherlands. All content represents
the opinion of the authors, which is not necessarily shared or endorsed by
their respective employers and/or sponsors. 
}
\thanks{The authors are with the Department of Cognitive Robotics (CoR), Delft University
of Technology, The Netherlands,
{\tt \{m.lodel; n.wilde; r.babuska; j.alonsomora\}@tudelft.nl}. R.~Babu\v{s}ka is also with CIIRC, Czech Technical University in Prague, Czech Republic.}%
}

\newcommand{\revision}[1]{#1}

\makeatletter
\newcommand{\multiline}[1]{%
  \begin{tabularx}{\dimexpr\linewidth-\ALG@thistlm}[t]{@{}X@{}}
    #1
  \end{tabularx}
}
\makeatother

\usepackage{tikz}

\newcommand{\copyrightnotice}{
\begin{tikzpicture}[remember picture,overlay]
    \node[anchor=south, yshift=0.5cm] at (current page.south) {
        \setlength{\fboxrule}{0pt}
        \fbox{\parbox{\dimexpr\textwidth-\fboxsep-\fboxrule\relax}{
            \scriptsize \color{gray} \centering © 2026 IEEE. Personal use of this material is permitted. Permission from IEEE must be obtained for all other uses, in any current or future media, including reprinting/republishing this material for advertising or promotional purposes, creating new collective works, for resale or redistribution to servers or lists, or reuse of any copyrighted component of this work in other works.
        }}
    };
\end{tikzpicture}
}

\begin{document}
\maketitle
\copyrightnotice
\thispagestyle{empty} 
\pagestyle{empty}

\begin{abstract}

The use of semantic features can improve the efficiency of target search in unknown environments for robotic search and rescue missions.
Current target search methods rely on training with large datasets of similar domains, which limits the adaptability to diverse environments. 
However, human experts possess high-level knowledge about semantic relationships necessary to effectively guide a robot during target search missions in diverse and previously unseen environments. 
In this paper, we propose a target search method that leverages expert input to train a model of semantic priorities. 
By employing the learned priorities in a frontier exploration planner using combinatorial optimization, our approach achieves efficient target search driven by semantic features while ensuring robustness and complete coverage.
The proposed semantic priority model is trained with several synthetic datasets of simulated expert guidance for target search.
Simulation tests in previously unseen environments show that our method consistently achieves faster target recovery than a coverage-driven exploration planner.

\end{abstract}

\section{Introduction}

Autonomous robots that can 
explore unknown environments efficiently by searching for objects of interest (OOI)
are promising tools 
in applications such as search and rescue, inspection, and environmental monitoring.
Efficient search typically relies on reasoning about semantic information in the scene and consequently determining \textit{where to search} next.
For example, search and rescue in an industrial site likely focuses on zones frequently used by workers, such as offices and storage rooms, that can be identified by characteristic objects like desks or shelfs.

By leveraging semantic priors of typical object arrangement,
recent works
 \cite{Chaplot2020, kim2022, ramakrishnan2022a, hahn2021,majumdar2022a, chenHowNotTrain2023,yokoyamaVLFMVisionLanguageFrontier2024,gintingSEEKSemanticReasoning2024}
have demonstrated this semantic exploration paradigm and achieved effective autonomous search behavior.
However, these methods either train on large domain-specific datasets \cite{xia2018a,chang2017} 
or use foundation models trained on internet-scale datasets, leading to common-sense reasoning capabilities \cite{majumdar2022a, chenHowNotTrain2023,yokoyamaVLFMVisionLanguageFrontier2024}.

However, domain-specific training data may not always be available for highly unpredictable and specific environments.
Moreover, foundation models require extensive computational resources that are infeasible for onboard deployment. \looseness=-1
On the contrary, pure coverage exploration methods \cite{cao2021, zhou2021a, huangFAELFastAutonomous2023} 
that effectively search everywhere can be deployed independently of domain priors but
can take a long time to find OOIs.

Specialized human operators, such as first responders in search and rescue, often have high-level knowledge about promising search locations based on observed semantic features, such as relevant objects.
However, increasing autonomy in exploration can be prefereable over teleoperation, as it reduces the operator's workload and is less reliant on robust communication.
Hence, we aim to leverage expert inputs to learn semantic priors for autonomous target search.
Independent of semantic information, 
coverage-driven exploration methods \cite{cao2021, zhou2021a,huangFAELFastAutonomous2023} can guarantee target discovery.
Therefore, 
a reliable semantic search approach must ensure that efficient exploration of the entire environment continues independently of semantic features and their learned priorities. 
\looseness=-1

\begin{figure}[t]
    \centering
    \includegraphics[width=\columnwidth, trim = 0 25 0 0, clip]{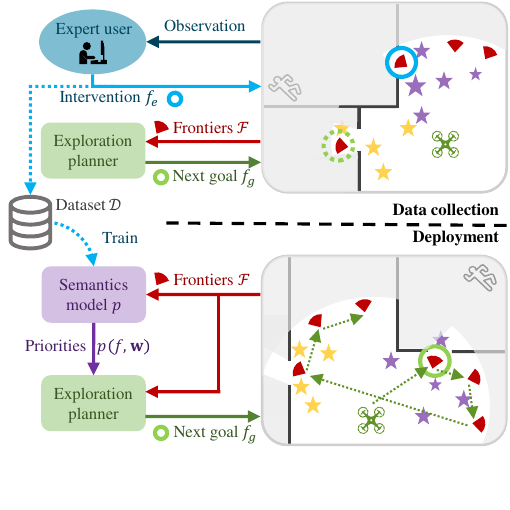}
    \caption{Conceptual overview of the proposed framework. During data collection, an expert generates interventions into the planner's goal output, prioritizing certain semantically relevant objects (depicted by stars). These are used to train a semantic model, which outputs priorities for each exploration frontier that, in turn, guide the exploration planner. The exploration planner outputs the next frontier viewpoint to navigate to.
    }
    \label{fig:overview}
    \vspace{-1.0em}
\end{figure}
In this paper, we present a
hierarchical exploration framework (\cref{fig:overview}), that can
learn semantics target search strategies from expert inputs.
Our paper makes the following contributions:

\begin{itemize}

\item We introduce a framework for learning a semantic priority function
that models the knowledge 
driving expert interventions, instead of imitating the expert.

\item We present a novel exploration planner leveraging these priorities
to prioritize promising frontiers.
\end{itemize}

In simulation experiments, we show that our framework achieves more efficient target search than coverage exploration after learning from only a small set of expert interventions. 
Moreover, our approach exhibits robust target search performance when learning from different simulated expert behaviors.

\section{Related Works}

In this section, we discuss existing approaches and how they relate to our work, focussing on semantic target search, learning human objective functions and coverage exploration. \looseness=-1

\subsubsection{Semantically-informed Target Search}

Semantically informed target search exploits environmental semantic features to accelerate target localization.
Several works address object search in unknown environments by learning semantic object relations from large-scale datasets \cite{xia2018a, chang2017}. Reinforcement learning (RL) approaches \cite{Chaplot2020, kim2022} train target search policies directly in simulation, whereas other methods \cite{ramakrishnan2022a, hahn2021} predict the cost-to-go of different positions, demonstrating better data efficiency than RL.
Conversely, zero-shot object search \cite{majumdar2022a, chenHowNotTrain2023,yokoyamaVLFMVisionLanguageFrontier2024} show that foundation models trained on internet-scale data can be used to predict likely object locations from semantic context in common indoor environments.
The authors of \cite{gintingSEEKSemanticReasoning2024} distill semantic knowledge from a large language model (LLM) into a smaller model for online inference of target probabilities.
In our paper, 
we learn a model of semantic priorities, similar to the prediction approaches \cite{ramakrishnan2022a, hahn2021}. We learn semantic knowledge from expert inputs 
unlike prior work using environment data
\cite{Chaplot2020, kim2022, ramakrishnan2022a, hahn2021}, comparable to distilling LLM common sense reason in \cite{gintingSEEKSemanticReasoning2024}.
\looseness=-1

\subsubsection{Learning Human Objectives}

Learning a priority model from human feedback involves learning the human's objective function.
In most works, this is formalized as learning a reward function \cite{christiano2017, abramson2022, wildeActivePreferenceLearning2020a} or action-value function \cite{zeng2021, spencer2022}.
\textit{Offline} feedback methods \cite{christiano2017, wildeActivePreferenceLearning2020a, zeng2021} query the human
for choice of different precomputed system behaviors \cite{christiano2017, wildeActivePreferenceLearning2020a}
or with states requiring a goal demonstration \cite{zeng2021}.
However, generating such queries is challenging in uncertain long-horizon tasks like exploration.
With \textit{online} feedback \cite{abramson2022, spencer2022}, the human chooses when to provide inputs as he interacts with an agent executing some baseline behavior.
Such online inputs can be binary feedback \cite{abramson2022}
or interventions with low-level demonstrations \cite{spencer2022}.
Our method considers online feedback in the form of expert interventions, similar to \cite{spencer2022}, demonstrating the preferred exploration frontiers.
We propose to learn an exploration priority model of different frontiers, similar to learning a value function over planning goals \cite{zeng2021}.
Moreover, we employ a stochastic model of expert actions, as in \cite{christiano2017}. \looseness=-1

\subsubsection{Coverage-driven exploration}

Coverage-driven exploration methods maximize the expected area coverage in order to build an occupancy map without considering semantic features.
Recently proposed methods employ 
combinatorial planning to visit all exploration frontiers \cite{cao2021, zhou2021a, meng2017,huangFAELFastAutonomous2023}, or navigation policies trained to maximize future coverage rewards using RL \cite{Niroui2019, lodel2022}.
Combinatorial planners repeatedly compute tours over all frontiers, allowing reasoning over long horizons and efficient navigation across frontiers.
These approaches have proven to work robustly in challenging real-world experiments \cite{cao2021, zhou2021a, meng2017,huangFAELFastAutonomous2023}.
We build on this concept and employ a combinatorial planner over frontiers, but consider semantic features for target search in the planner.
To this end, we propose a planner formulation that, similar to \cite{huangFAELFastAutonomous2023}, 
can schedule frontiers based on a priority measure
but prioritizes based on both semantics and coverage. \looseness=-1

\section{Problem Formulation}
\label{sec:problem_form}

We consider the usecase where an autonomous robot searches for a target object in an unknown environment $\mathcal{W} \subset \mathbb{R}^2$ with obstacle-free space $\cW_f \subset \cW$.
The robot's position at time $t$ is denoted by $\mathbf{x}_t \in \cW_f$, and it starts exploring from an initial position $\vx_0$.
The robot moves incrementally with actions $\va \in \mathbb{R}^2$ bounded by $| \va | < \delta_\text{max}$, where $\delta_\text{max}$ is the maximum distance per time step.
An action $\va$ can only be applied if the new position is in free space, i.e., $\vx_t + \va \in \cW_f$.
A more complex dynamic model can be considered, for example, by tracking the reference $\va$ with a model predictive controller, as in \cite{lodel2022}.

\emph{Occupancy Map}
From range observations with sensing range $r$ until time $t$, the robot builds an occupancy map $\vM_t$ of the environment. 
The occupancy map is represented as a grid, where cells correspond to evenly spaced positions $\vx \in \cM,\ \cM \subset \cW$, and are in one of three discrete states: unexplored (0), free (1), or occupied by obstacles (-1), i.e., $\vM \in \{-1,0,1\}^{m \times m}$ with grid size $m$. \looseness=-1

\emph{Semantic features:} The robot observes objects of different semantic classes $\cS$ when exploring the environment. An object is denoted as $o = (o^p, o^s)$, defined by its position $o^p \in \cW_f$ and its semantic class $o^s \in \cS$.
We assume that the robot's sensors can detect objects around the robot within radius $r$ that are not occluded by obstacles. The set $\cO_t$ denotes the objects observed up to time t.
We further assume that semantic relationships between objects of different classes exist, i.e., the presence of certain objects can indicate an increased or reduced likelihood of other objects being present close by. %

\emph{Expert input:} We further assume the availability of an expert with knowledge of semantic relationships relevant to the search task.
Leveraging this knowledge, the expert can infer likely target locations from the
observed objects $\cO_t$, 
and
guide the robot to the target with waypoint inputs $\vh \in \cW_f$ that follow some expert policy $\mu$, i.e., $\vh_t \sim \mu(\vh_t | \cO_t)$.
From this expert interaction, a dataset $\cD$ of 
expert inputs $\vh_t$ with associated observations $\vM_t, \cO_t$ is recorded.

\emph{Problem:} given a dataset of human-guided target search trajectories $\cD$,
the problem is to find a navigation policy $\pi_\cD$
controlling the robot with $\va_t = \pi_\cD(\vM_t, \cO_t)$,
that minimizes the distance traveled until discovering a target object $o_g$,
using the map and object memory as current knowledge about the environment. With 
$H$ as the final time step, the problem is formulated as
\begin{align}
    \pi_\cD =
    \arg\min_{\pi} &
    \sum_{t=0}^{H-1}
    \| \vx_{t+1} - \vx_{t} \|
    \nonumber \\
    \label{eq:problem}
    \text{s.t.} \quad& \|o_g^p - \vx_H \| \leq r
    \\
     & \vx_{t} = \vx_{t-1} + \pi(\vM_t, \cO_t),\ 
     \forall t \in \{1,\dotsc,H\},
     \nonumber
\end{align}
where the first constraint indicates target discovery at time $H$. \looseness=-1

\section{Method}
\label{sec:method}

Exploring an unknown environment to search for a target object requires continually solving two subproblems: Semantic scene understanding, or \textit{where is it promising to explore}, and planning, or \textit{where to go next},
given a set of regions to explore.
We choose to solve these two problems in a hierarchical framework depicted in \cref{fig:overview} to obtain a data-efficient approach 
robust to unseen scenarios.
\looseness=-1

Both subproblems are solved using the concept of frontier exploration \cite{yamauchi1997}, which we formalize for our method in \cref{sec:frontier_exploration}.
The first problem of semantic scene understanding is formalized as evaluating different frontiers with a semantic priority function. Specifically, we present an approach to learning such a semantic priority function from expert interventions in \cref{sec:method_model}.
To solve the second problem of efficient navigation, we devise a combinatorial target search planner leveraging the learned semantic priority function. Specifically, the planner determines a visitation order such that semantically promising frontiers with increased probability of target discovery are prioritized, thus approximately solving Problem \eqref{eq:problem}. \looseness=-1

\subsection{Frontier Exploration}
\label{sec:frontier_exploration}

In this section, we describe how our approach formalizes the concept of frontier exploration, drawing inspiration from recent works \cite{zhou2021a,huangFAELFastAutonomous2023}.
Frontiers are the boundaries between explored and unexplored space in $\vM_t$ and are used to derive a discrete set of candidate
positions for observing unexplored space, called
\textit{frontier viewpoints}, that enable efficient exploration planning.
To obtain such frontier viewpoints $f \in \cF_t,\ \cF_t \subset \cW_f$ and efficient paths between them, a topological graph $\cG_t = (\cV_t, \cE_t)$ is gradually constructed
in the free space of $\vM_t$.
At every timestep, the graph is expanded using a sampling-based method from \cite{huangFAELFastAutonomous2023}, ensuring sparsity.
We consider every node $v_i \in \cV_t$ as a potential frontier viewpoint if sufficient unexplored area is visible from $v_i$.
To this end,
we define a \textit{coverage gain} function $\cI(v_i) \colon \cV \mapsto \mathbb{R}$ that denotes the gain in map coverage when observing frontiers from $v_i$.
Specifically, the coverage gain approximates the expected gain in the covered area by casting a fixed number of equally spaced rays from $v_i$ and averaging the number of visible unexplored cells on each ray.
The set of frontier viewpoint nodes $\cF_t$ are those with $\cI(v_i) > \cI_\text{thres}$, referred to as \textit{frontiers} $f \in \cF_t$.
We further assume that $\vM_t$ is clustered into regions, e.g., rooms in a building, using a method such as \cite{hughes2022}, and 
each frontier is associated with a region. \looseness=-1

\subsection{Modeling Expert Frontier Choices}
\label{sec:method_model}

This section formulates a model of expert behavior 
that will be used to train the semantic priority model.
When collecting data, the expert can \textit{intervene} in the robot's exploration behavior at any time $t$ by determining the next \textit{waypoint} that the robot will navigate to. \looseness=-1

\subsubsection{Semantic Priority Function}
The expert considers each available frontier $f \in \cF_t$ as a potential intervention waypoint,
and evaluates how likely exploring a frontier $f$
leads to the target object,
based on nearby objects and the expert's semantic knowledge.
This evaluation is formalized
as a semantic priority function $p(f, \vw)$. We model this function with a weighted sum where $\vw \in [0,1]^n$ are weights on $n$ different features, such as semantic classes.
These features form a semantic feature vector $\vphi(f)$ for each frontier $f$.
Thus, the priority function can be written as, 
\begin{equation}
\label{eq:method_model_linear}
    p(f, \vw) = \vw^T \vphi(f)
\end{equation}
which is common in preference learning \cite{wildeActivePreferenceLearning2020a,sadighActivePreferenceBasedLearning2017} to allow learning from a small number of expert interventions.
The weight vector $\vw$ used by the expert is unknown and will be estimated from expert inputs.
\looseness=-1

\subsubsection{Semantic Feature Vector}
The feature vector $\vphi(f)$ consists of two parts: Semantic features $\vphi_s$ and an auxiliary region novelty feature $\phi_n$, i.e., $\vphi(f) = [\vphi_s(f),\ \phi_n(f)]^T$.
Semantic features $\vphi_s$ describe the occurrence of different semantic classes around the frontier node $f$. 
Each semantic feature needs to capture the presence of the semantic class in the vicinity and in the region of the frontier. 
Both effects are part of the semantic feature vector: A binary \textit{local} semantic vector 
$\vphi_{s,l}(f) \in \{0,1\}^{|\cS|}$ 
indicating if a class is visible within a small radius around $f$, and a binary \textit{region} semantic vector 
$\vphi_{s,r}(f) \in \{0,1\}^{|\cS|}$ 
indicating if a class is present in the same region as $f$. We combine both as $\vphi_s(f) = \lambda \vphi_{s,r}(f) + (1-\lambda) \vphi_{s,l}(f)$
with $\lambda$ as hyperparameter.
\revision{Aggregating semantic features in regions allows the search process to exploit the fact that objects are not distributed randomly, but are clustered within functional areas that serve as strong indicators for related objects.}
The region novelty feature $\phi_n$
captures the expert's interest in observing semantic information in unexplored regions,
and remains 1 unless a small number of objects are observed in the region of frontier $f$. \looseness=-1

\subsubsection{Expert Intervention Model}
\label{sec:method_model_feedback}
When providing online waypoint interventions, the expert's capability to quickly plan over multiple frontiers 
is limited. 
Hence, we model the expert behavior with a greedy algorithm for choosing the next frontier.
This greedy choice 
is
modeled by a utility function $u(f)$ assigned to each frontier $f \in \cF_t$.
The expert is also interested in coverage exploration to guarantee search success without relying only on semantic priorities.
Furthermore, the expert aims at minimizing the traveled distance until discovering the target object (Problem \eqref{eq:problem}), which is modeled by discounting
the semantic priority by the traveling costs to the frontier.
We propose a greedy choice model, maximizing
a utility $u(f, \vw)$,
that combines the semantic priority $p(f,w)$ with the coverage gain $\cI(f)$ and the distance to the frontier, given by \looseness=-1
\begin{equation}
    \label{eq:method_model_utility}
    u(f, \vw, w_\cI) = \delta (f) \bigl(p(f, \vw) + w_\cI \cI(f) \bigr).
\end{equation}
Here $\delta (f)$ is the distance-based discounting function, defined as
$\delta (f) = 1 - (\nicefrac{d_t(f)}{\max_{f' \in \cF_t} d_t(f')}) + \epsilon$, with $d_t(f)$ expressing the traveling distance from the current position $\vx_t$ to $f$ through $\cG_t$ and $\epsilon$ defining the minimum discounting factor.
The utility model in \cref{eq:method_model_utility} adds a coverage term weighted by the learnable parameter $w_\cI$ to the semantic priority $p$ and discounts this extended priority
by a factor $\delta (f)$ decreasing with distance to the frontier.
Normalizing distances in $\delta (f)$ ensures consistent utility values across different frontier sets $\cF$.
\newcommand{\tvphi}{\wt{\vphi}}
\newcommand{\tvw}{\widetilde{\vw}}
Finally, the utility function can be written as a linear model $u(f, \vw, w_\cI)=u(f, \tvw)=\tvw^T \tvphi(f)$ with augmented weights $\tvw = [\vw, w_\cI]^T$ and features $\tvphi(f)=\delta(f) \bigl[ \vphi(f),\ \cI(f) \bigr]^T$. \looseness=-1

\subsubsection{Pairwise Choice Model}
\label{sec:method_model_pairwise}

Next, we derive a probabilistic model of the expert's frontier choice to learn the expert weights from noisy expert intervention data.
We model the expert preference for a frontier $f_e \in \cF_t$ as pairwise choices between $f_e$ and all other available frontiers. 
Hence, the expert prefers frontier $f_e$ if its utility is higher than of all other available frontiers,
i.e., if 
$u(f_e, \tvw) \geq u(f, \tvw),\ \forall f \in \cF_t \setminus \{f_e\}$.
The Bradley-Terry model \cite{bradleyRankAnalysisIncomplete1952,christiano2017}
defines the probability of choosing $f_i$ over $f_j$, denoted by $\mathbb{P}(f_i \succ f_j)$, 
as a logistic sigmoid function $\sigma$ of their utility difference, 
i.e., $\mathbb{P}(f_i \succ f_j) = \sigma(\beta (u(f_i) - u(f_j)))$. 
Here, $\beta$ is the rationality parameter modeling uncertainty in the expert's decision-making process. 
However, this model assumes that probabilities converge to 0 or 1 for large utility differences.
We choose to modify this model to account for a residual error probability independent of the utility difference and $\beta$, considering cases where the utility model cannot capture potentially complex expert reasoning.
Inspired by \cite{wildeActivePreferenceLearning2020a}, we define $\rho \in[0,0.5]$ as a lower bound on the probability of wrong choice independent of the utilities, used to formulate a scaled and shifted sigmoid function $\sigma_\rho$:
\begin{equation}
    \sigma_\rho(x) = (1 - 2 \rho) \sigma(x) + \rho.
\end{equation}
Then, the probability that the expert chooses $f_e$ over any $f \in \cF_t \setminus \{f_e\}$, given weights $\tvw$, is modeled as
\begin{equation}
    \mathbb{P}(f_e \succ f|\tvw) = 
    \sigma_\rho
    \bigr(
        \beta\ \tvw^T (\tvphi(f_e) - \tvphi(f))
    \bigl).
    \label{eq:expert_binomial}
\end{equation}
Here, $\beta$ and $\rho$ are tunable hyperparameters.
This proposed model captures noisy expert waypoint interventions based on the semantic priority function $p(f,\vw)$.

\subsubsection{Learning Expert Weights}
The final step of the expert model is learning the expert weights from recorded intervention data.
Given a set of $N$ choices 
$\cC = \{(f^1_e, f^1),\ldots, (f^N_e, f^N)\}$
from the expert and assuming a uniform prior, 
we obtain the maximum likelihood estimate of the expert weights given the expert choices using gradient-based optimization, solving
\begin{equation}
    \label{eq:method_model_optimization}
    \tvw_{mle} = \argmin_{\tvw} \sum_{(f_e,f) \in \cC} \bigl[ -\log \mathbb{P}(f_e>f|\tvw) \bigr],
\end{equation}

\newcommand{\hp}{\hat{p}}

\subsection{Frontier Planning for Priority-Aware Exploration}
\label{sec:method_planner}

In this section, we introduce a global planning method for target search given a semantic priority model (\cref{sec:method_model}). 

\subsubsection{Target Search as Combinatorial Optimization}
We extend coverage-maximizing exploration methods that leverage combinatorial planning over frontier viewpoints \cite{cao2021, zhou2021a, meng2017,huangFAELFastAutonomous2023}, 
by incorporating semantic priorities.
The combinatorial planner generates a visitation order, or \textit{tour}, through all known frontier viewpoints. 
For effective target search, promising frontiers should be scheduled earlier in the tour, such that the distance to the target object is minimized (\cref{eq:problem}).
Consequently, we need to minimize the total distance traveled to frontiers with high semantic priority values $p(f, \vw)$, which are expected to be close to the target. 
We frame target search as a variant of the Minimum Latency Problem (MLP) \cite{blumMinimumLatencyProblem1994}, denoted as weighted MLP (WMLP), 
where the planned visitation latencies of the frontiers are weighted using the learned semantic priority model $p(f, \vw)$.

\subsubsection{Planner Formulation}
We formulate the planning problem over
a subset of nodes in the topological graph $\cG_t$
composed of the
the frontier nodes $\cF_t$ and the robot's current node $v_t \in \cV_t$,
denoted as $\cF_t' = \cF_t \cup \{v_t\}$.
A distance matrix $D$ contains the lengths of the shortest paths through $\cG_t$ between all pairs of nodes in $\cF_t'$.
The tour $T$ is a sequence of all nodes in $\cF_t'$ 
describing the planned visitation order, always starting with the robot node $v_t$.
We denote that frontier node $f_i$ is scheduled at position $j$ in the tour as $T(j) = f_i$ for $j>0$,
while $T(0) = v_t$.
Let $P(f)$ be a priority function that assigns each node in $\cF_t'$ a priority weight,
and $m=|\cF_t'|$, then the WMLP objective is
\begin{equation}
\label{eq:planner_cost}
    \min_{T}\
    \sum_{i=1}^{m-1}
    P(T(i))
    \sum_{j=1}^{i} D(T(j-1), T(j)).
\end{equation}
Assuming a unit velocity, this problem minimizes a priority-weighted sum of the visitation latencies of each frontier, favoring earlier visits to high-priority frontiers.
The priority function $P(f)$ leverages the learned semantic priorities $p(f, \vw_{mle})$ to prioritize regions that likely lead to the target. 
Combining semantic priorities with expected coverage gain ensures robust exploration when the semantic priorities are ambiguous or incorrect, e.g., when encountering unseen states. 
Instead of the weighted sum model used in \cref{eq:method_model_utility}, we propose a heuristic priority function $P(f)$ that always pursues coverage but is biased to semantically important frontiers, which we found more robust for the WMLP planner.
Let $p'(f)=\nicefrac{p(f)}{p_{max,t}}$ 
be the normalized semantic priority 
of frontier $f$ with $p_{max} = \max_{f \in \cF_t} p(f)$, 
then $P(f)$ is given by
\begin{equation}
    \label{eq:planner_priority}
    P(f)
    =
    \left(p'(f, \vw_{mle}) + \alpha \right) \cdot \cI(f).
\end{equation}
Here, $\alpha \in [0,1]$ is a hyperparameter controlling the trade-off between semantic priority and coverage gain.
Note that while we learn the weight vector $\tvw_{mle} = [\vw_{mle}, w_{\cI,mle}]^T$, we only use $\vw_{mle}$ for inferring frontier priorities, and discard the learned weight $w_{\cI,mle}$ of the coverage gain.
This allows for tuning the balance between target search and coverage to reflect confidence in the learned semantic priority.
The normalization of $p(f)$ addresses states where an important frontier only has a single non-zero feature or low feature activations in $\vphi(f)$, which can lead to a low-valued semantic priority $p(f)$. 
By normalizing $p(f)$ by the maximum value in the current state, the combination of semantic priorities and coverage gains proposed in \cref{eq:planner_priority}, with a fixed $\alpha$ across different scenarios, becomes more robust.
\begin{algorithm}[t]
    \caption{Prioritized exploration planning}
    \label{alg:planner}
    \SetCommentSty{slshape}
    \SetKwInOut{Input}{Input}
    \SetKwComment{Comment}{$\triangleright$\ }{}
    \Input{Semantic priority model weights $\vw_{mle}$}
    Init $\mathcal{G}_t \gets \emptyset$, $\cF_t \gets \emptyset$, and unexplored map $\vM$
    \\
    \ForEach{time step $t$ from 1 until $t_\mathrm{end}$}
    {
        $\vM_t,\ \cG_t,\ \cF_t,\ v_t \gets$ \textsc{PerceptionUpdate}()
        \\
        \If{Target found \textbf{or} $\cF_t=\emptyset$}
        {
            \textbf{break}
        }
        \If
        {\label{alg:tsp_condition}
            $\cF_t \neq \cF_{t-1}$
            \textbf{or}
            $\cI(f)$ changed for any $f \in \cF_t$
        }
        {
            $\vP \gets$  \textsc{FrontierPriorities}$(\cF_t, \vw_{mle})$
            \Comment*[f]{Computes vector with \cref{eq:planner_priority}~$\forall f \in \cF_t$}
            \\
            $T \gets$ \textsc{LNSsolver} ($\mathcal{G}_t,\ \cF_t,\ v_t,\ \vP$)
            \\
            $f_{g} \gets T(1)$
            \Comment*[f]{Set goal node to next in tour}
        }
        \Else
        {
            \If{
                $v_t = f_{g}$
                \label{alg:tsp_condition_else}
            }
            {
                $f_{g} \gets$ next frontier in $T$
            }
        }
        $\cP \gets$ \textsc{ShortestPath}$(\cG_t, v_t, f_{g})$
        \\
        Move to next vertex in $\cP$
    }
\end{algorithm}

\subsubsection{Plan Execution and Control}
We now explain how the exploration planner navigates the robot through the environment, which is summarized in \cref{alg:planner}.
At every time step, the perception module updates the topological graph, the frontier set, and the robot's position.
The tour is replanned whenever the current frontier set $\cF_t$ or their coverage gains change (\cref{alg:tsp_condition}).
In that case, the priorities of all current frontiers are updated, and then a new tour $T$ is found by minimizing \cref{eq:planner_cost} using a large neighborhood search (LNS) algorithm \cite{pisingerLargeNeighborhoodSearch2019}.
In each iteration, our custom LNS algorithm uses random destruction of up to 30\% of the tour and reconstructs it using the
cheapest insertion heuristic \cite{rosenkrantz1977} follwed by a 2-opt swapping search \cite{croesMethodSolvingTravelingSalesman1958}.
Given a new tour, the next frontier in the tour is chosen as the subgoal $f_g = T(1)$.
If the tour is not recomputed and the subgoal $f_g$ has been reached (\cref{alg:tsp_condition_else}), the next node in $T$ is set as the goal. Otherwise, $f_g$ stays the same.
The shortest path to $f_g$ is planned using A$^*$ \cite{hart1968} through $\cG_t$, and the robot moves to the first node in the path 
$v_{p,1} \in \cV_t$, 
applying $\va = \| v_{p,1} - v_t \|$.

Under the assumption of a perfect perception module that will correctly detect all frontiers within its range, our planning approach will eventually visit every frontier becoming available, independent of the priority function.
Since only graph nodes with a minimum coverage gain are considered frontier viewpoints, tours will not include already visited frontiers, guaranteeing that the robot always moves towards unexplored spaces.
Therefore, our planner can ensure complete exploration of the environment.

\section{Experiments}
\label{sec:experiments}

\subsection{Experimental Setup}
\label{sec:exp_setup}

Experiments are conducted in a Python-based 2D simulator with simplified sensing and navigation \cite{everett2018,lodel2022}.
Important aspects of the experiments are detailed below.

\subsubsection{Scenario Setup}

We use ProcThor \cite{deitkeProcTHORLargeScaleEmbodied2022} to sample multi-room indoor floorplans and realistic object placements with 4 different room categories (kitchen, bathroom, living room, bedroom).
We generate environments with 3 kitchens, 3 bathrooms, 1 living room, and 1 bedroom, arranged with constrained connectivity
(bedroom only accessible from the living room, bathrooms to the living room via the kitchens). 
We configure two scenario setups with a different target object and starting room type, detailed in the following sections.
Top-down maps and object data are extracted for the simulator, and additional small objects are sampled to increase semantic feature density.
Scenarios are curated to ensure challenging tasks where semantic features offer an advantage for target search.
Both setups use 30 scenarios for generating intervention datasets and 34 scenarios for obtaining the evaluation results. \looseness=-1

\subsubsection{Oracle-based Data Generation}
\label{sec:exp_setup_data}
For training the semantic priority model, we generate synthetic interaction datasets by simulating expert interventions with an oracle model based on the expert model in \cref{sec:method_model}. 
\revision{
The oracle's priority model was designed and tuned to approximate a moderately rational expert providing occasional guidance while still allowing exploration. 
}
It assigns priorities based on room types, favoring frontiers in target rooms or exploring unseen rooms.
Rooms are classified using a list of characteristic object classes for each room type.
The expert model parameters (\cref{sec:method_model}) used by the oracle are $\beta=25.0$, $\rho=0.0$, $\epsilon=0.2$, and an intervention threshold $\tau=0.05$.
We also generate a dataset with an exponential distance discounting model, that is $\delta(f) = \exp{(-\gamma d_t'(f))}$ with $\gamma=0.1$, as well as datasets with varied $\beta$ and $\tau$
to evaluate the robustness of our method to different expert behaviors (see \cref{sec:exp_robustness}).
Finally, we vary the number of episodes $N_{eps}$ in the intervention dataset to evaluate the data efficiency of our method. \looseness=-1

\subsubsection{Training}
The weights of the semantic priority model are trained using Adam \cite{kingmaAdamMethodStochastic2015}
minimizing the negative log-likelihood of the observed expert choices (\cref{eq:method_model_optimization}) for 2000 epochs with learning rate 0.01. 
For each dataset, training uses 10 different random seeds, and the fixed expert model parameters are $\beta=10.0,\ \rho=0.1,\ \lambda=0.7$.

\subsection{Overview of Experiments}
\label{sec:exp_overview}

We evaluate the performance of our method in two task setups (see \cref{sec:exp_setup}) and present both qualitative and quantitative results.
A coverage baseline, similar to \cite{huangFAELFastAutonomous2023},
uses the planner proposed in \cref{sec:method_planner},
but with the priority function $P(f) = I(f)$.
For both task setups, we first present qualitative results to illustrate an example scenario and the behavior of our method and the baseline.
Second, we evaluate the target search performance of our method using quantitative metrics and compare it to different oracle methods, serving as upper bounds for the search performance. 
Using the same metrics,
we evaluate the robustness of our method to different expert datasets by varying the number of interventions and parameters of the oracle model in the first scenario setup.
\looseness=-1

\subsubsection{Metrics}
\label{sec:exp_overview_metric}
We evaluate target search performance using the following metrics:
\begin{itemize}
    \item \textit{Path Length Ratio to Coverage (PLR)}: The episode-wise ratio of the path lengths $l$ until target discovery between the \revision{compared semantic method $l_{\text{sem}}$ and the coverage planner $l_{\text{cov}}$}, i.e., $\text{PLR} = l_{\text{sem}}/l_{\text{cov}}$. 
    The compared method reaches the target faster than coverage exploration for $PLR<1$.
    \item \textit{Success weighted by Path Length (SPL)}: The ratio of the traveled and shortest path to the target. A value of 1 indicates the shortest possible path to the target.
\end{itemize}
While the SPL metric is common in object search \cite{anderson2018}, the PLR metric is proposed as the main metric to evaluate the efficiency of our method compared to the coverage planner
as it quantifies the relative advantage over coverage per scenario. \looseness=-1

\subsubsection{Oracle Methods}
In our performance evaluation, we compare our method to the following oracle methods:
\begin{itemize}
    \item \textit{Oracle Interventions} waypoint interventions from the oracle model overwrite the coverage baseline behavior
    \item \textit{Oracle Priorities} guides the planner with the semantic priorities from the oracle model.
    \item \textit{Linear Oracle} uses a linear expert model (as \cref{eq:method_model_linear} with hand-tuned weights to obtain semantic priorities.
\end{itemize}

\subsection{Primary Scenario Results}

In the primary scenario setup, the target object is a bed in the bedroom, and the robot is initialized in one of the kitchens. 
Therefore locating the living room first and then the door to the bedroom is necessary. 

\begin{figure}
    \centering
    \begin{subfigure}{0.244\textwidth}
        \centering
    	\includegraphics[width=\linewidth, trim = 50 70 270 120, clip]{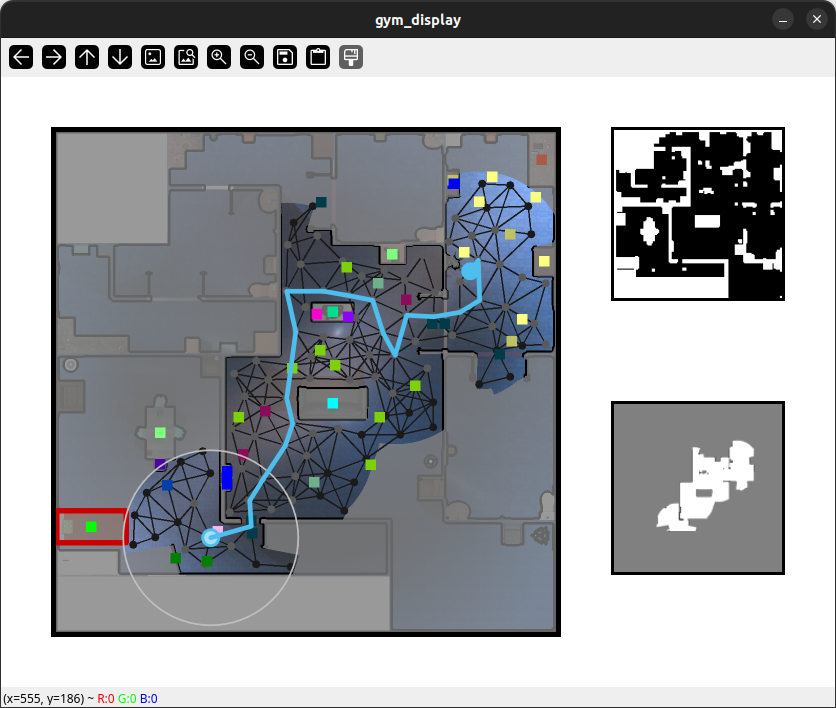}
    	\captionsetup{justification=centering}
    	\caption{Learned Priority Model\\$\text{SPL}=0.784,\ \text{PLR}=0.226$}
    	\label{fig:exp_result_model}
    \end{subfigure}%
    \begin{subfigure}{0.244\textwidth}
        \centering
    	\includegraphics[width=\linewidth, trim = 50 70 270 120, clip]{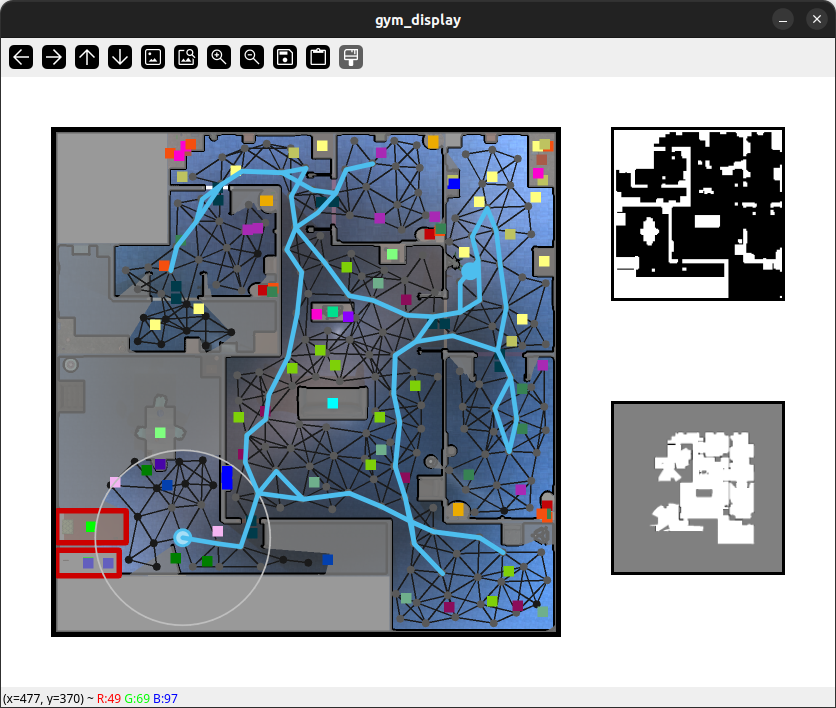}
    	\captionsetup{justification=centering}
    	\caption{Coverage Exploration\\$\text{SPL}=0.177,\ \text{PLR}=1.0$}
    	\label{fig:exp_result_cov}
    \end{subfigure}%
    \caption{
        Top-down views of an example scenario of the first task setup comparing coverage-driven exploration with our learned semantic priority model.
        Frontier nodes of the topological graph are colored black, and others are gray. Blue edges visualize the path taken by the robot; the larger blue circle is the robot's position at target discovery time, and the smaller blue circle is the initial position. The red rectangles are target objects. 
        Object instances are visualized as small squares colored according to semantic class. \looseness=-1
    }
    \label{fig:exp_qual_semantic}
    \vspace{-0.5em}
\end{figure}

\subsubsection{Qualitative Results}
\Cref{fig:exp_qual_semantic} compares the paths taken by the coverage planner and our target search planner with learned priorities in an example scenario (dataset $N_{eps}=30$).
The target object is in the bedroom (lower left) the robot starts in a kitchen (top right) 
and a large living room at the center connects the bedroom and kitchens. 
\Cref{fig:exp_qual_semantic} shows that our framework can guide the robot to the target object using a substantially shorter path than the coverage planner. 
Initially, the robot navigates to the living room instead of exploring the other doorway below, as 
observed objects in the living room are prioritized.
The robot discovers a higher density of relevant objects in the lower part of the living room in turns in that direction.
The robot also prioritizes door objects to search for the bedroom, leading the robot to the correct target room.
Finally, discovered bedroom objects yield the highest priority and lead the robot to the target object. 
Conversely, the coverage planner prioritizes frontiers only based on coverage gain and first explores the large open spaces in the living room and, subsequently, the smaller rooms, ignoring semantic features.
These exemplary results illustrate that our framework can leverage semantic features in the environment to achieve better target search efficiency than coverage-driven exploration. \looseness=-1

\subsubsection{Performance Results}
\label{sec:exp_performance}

\begin{figure}
    \centering
    \includegraphics[width=\columnwidth, trim=0 20 0 0]{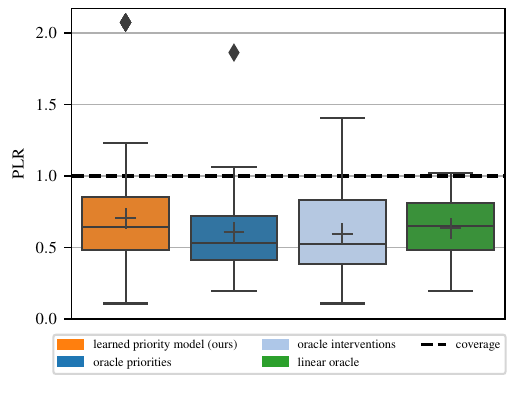}
    \caption{
        Performance results in the primary target search task: Comparison of our method to oracle methods, displaying the episode-wise path length ratio (PLR) to the coverage baseline (dashed line) as boxplots.
    }
    \label{fig:boxplot_main}
\end{figure}

We evaluate the target search performance of our method using $N_{eps}=30$ in multiple test scenarios. 
The 340 episode results (10 training seeds and 34 test scenarios) are visualized as boxplot in \cref{fig:boxplot_main}.
The orange boxplot shows that our method significantly outperforms the coverage planner (dashed line) in most scenarios (median $\text{PLR} = 0.644$), 
up to a best-case performance of $\text{PLR} = 0.11$.
In 88\% of episodes, our method is more efficient than the coverage planner, and in 97\% of the episodes, PLR is smaller than 1.3, indicating that cases where our method misguides the robot are rare.
Moreover, our approach matches the linear oracle and is only slightly outperformed by the non-linear oracle guidance.
These results show that our approach learned the underlying semantic priorities of the oracle expert and effectively leverages them in multiple unseen scenarios. 
That is, by incorporating the learned priorities in the cost function of the planner, it prioritizes exploration frontiers likely to lead to the target. \looseness=-1
\Cref{tab:results}
additionally reports the SPL metric indicating a strong advantage in absolute target search performance over coverage exploration and competitive performance compared to oracle methods.

\setlength{\tabcolsep}{4pt}
\begin{table}
    \centering
    \caption{Comparison of our method with the coverage baseline and oracle methods \revision{using the SPL metric defined in \Cref{sec:exp_overview_metric}, given as {mean~$\pm$~std}.}}
    \label{tab:results}
    \begin{tabularx}{\columnwidth} { 
        l
        >{\centering\arraybackslash}X 
        >{\centering\arraybackslash}X 
    }
    \toprule
    \textbf{Method}  & \textbf{SPL (Task Setup 1)} & \textbf{SPL (Task Setup 2)} 
    \\
    \midrule
    Coverage Priorities
    & 0.406 $\pm$ 0.196 & 0.341 $\pm$ 0.313 
    \\
    Oracle Priorities
    & 0.704 $\pm$ 0.202 & 0.564 $\pm$ 0.275 
    \\
    Oracle Intervention
    & 0.712 $\pm$ 0.206 & 0.529 $\pm$ 0.281 
    \\
    Linear Oracle Priorities 
    & 0.650 $\pm$ 0.207 & 0.520 $\pm$ 0.271 
    \\
    \textbf{Learned Priorities (ours)}
    & \textbf{0.627 $\pm$ 0.225} & \textbf{0.520 $\pm$ 0.313} 
    \\
    \bottomrule
    \end{tabularx}
    \vspace{-0.5em}
\end{table}

\looseness=-1

\subsubsection{Robustness to Data Variation}
\label{sec:exp_robustness}

\begin{figure}[t]
    \centering
    \includegraphics[width=\columnwidth, trim=0 0 0 0]{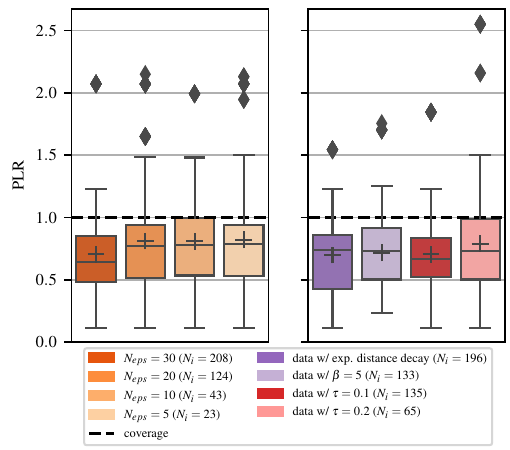}
    \caption{
        Comparison of different dataset sizes and oracle behaviors used for training, displaying PLR performance of the resulting priority models.
    }
    \label{fig:boxplot_ablation}
    \vspace{-0.5em}
\end{figure}

Next, we analyze the robustness of our method to different dataset sizes $N_{eps}$ and expert behavior by varying the oracle parameters.
For each dataset, semantic priority models are trained and tested as described in \cref{sec:exp_performance}.
\Cref{fig:boxplot_ablation} shows the resulting PLR boxplots.
The left subplot shows the results for a reduced number of training episodes, 
($N_{eps} = 30$ is the same as in \cref{fig:boxplot_main}). 
It is evident that with all 4 datasets, similar PLR performance is achieved. 
However, performance drops from $N_{eps} = 30$ to $N_{eps} = 20$, but further reduction up to $N_{eps} = 5$ does not affect the performance. 
Note that our method can achieve strong target search efficiency with only $N_i=23$ expert interventions ($N_{eps} = 5$). 
A substantial improvement with more training data is only observed at $N_{eps}=30$, which likely results from highly informative data points that only occur in this dataset, indicating that additional data can lead to further performance gains.
The right subplot shows the results for 4 different oracle variations: 
exponential distance discounting instead of linear (\cref{eq:method_model_utility}),
reduced expert rationality $\beta$ (increased noise, \cref{eq:expert_binomial}),
and increased expert intervention threshold (less engaged, more selective expert), all with $N_{eps} = 30$.
Our method is robust to these changes and yields similar results across all variations. 
The lowest performance occurs for $\tau=0.2$, since a less engaged expert might miss providing some informative interventions. \looseness=-1

\subsection{Secondary Scenario Results}
\label{sec:exp_secondary}

The secondary scenario setup uses the same maps as the primary, but the target object is a toilet in one of the three bathrooms, and the robot starts in the living room. 
Here, the robot must first prioritize finding any of the kitchens that will lead to the bathrooms and the target object. 
In this setup it is harder to leverage semantic features as two kitchen-bathroom pairs might attract the robot but do not yield the target. %

\begin{figure}[ht]
    \centering
    \begin{subfigure}{0.244\textwidth}
        \centering
    	\includegraphics[width=\linewidth, trim = 50 70 270 120, clip]{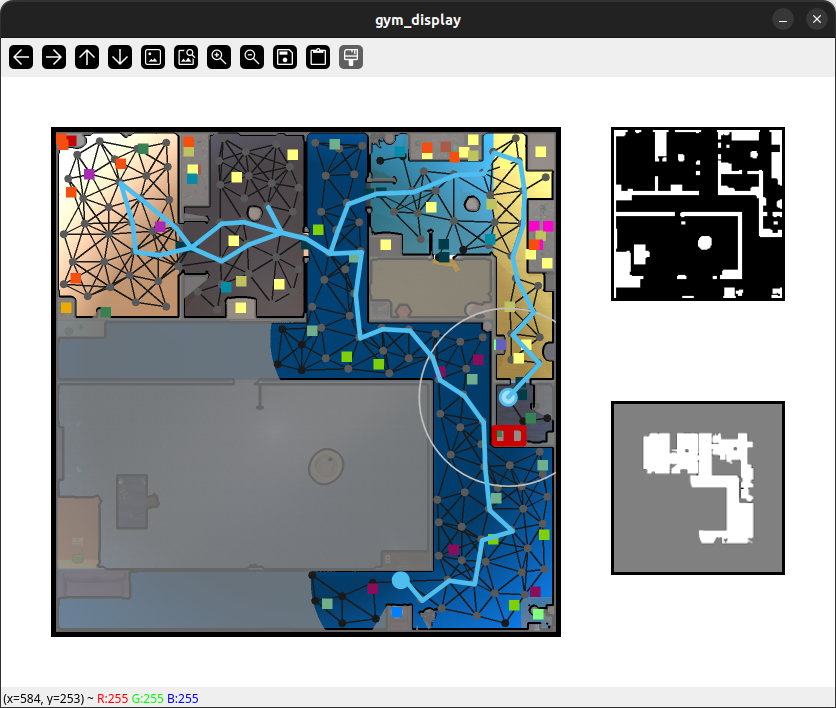}
    	\captionsetup{justification=centering}
    	\caption{Learned Priority Model\\$\text{SPL}=0.509,\ \text{PLR}=0.530$}
    	\label{fig:exp_secondary_qual_sem}
    \end{subfigure}%
    \begin{subfigure}{0.244\textwidth}
        \centering
    	\includegraphics[width=\linewidth, trim = 50 70 270 120, clip]{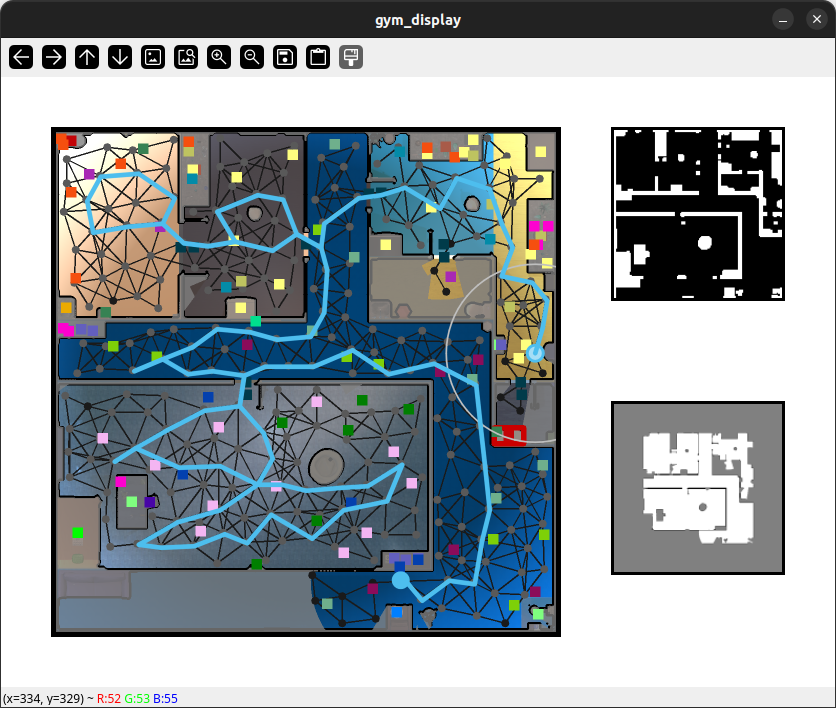}
    	\captionsetup{justification=centering}
    	\caption{Coverage Exploration\\$\text{SPL}=0.270,\ \text{PLR}=1.0$}
    	\label{fig:exp_secondary_qual_cov}
    \end{subfigure}%
    \caption{
        Top-down views of an example scenario of the second task setup comparing the behavior of coverage-driven exploration and our learned semantic priority model.
        Visuals follow the same conventions as in \cref{fig:exp_qual_semantic}.
    }
    \label{fig:exp_secondary_qual}
    \vspace{-0.5em}
\end{figure}

\begin{figure}[t]
    \centering
    \includegraphics[width=\columnwidth, trim=0 20 0 0]{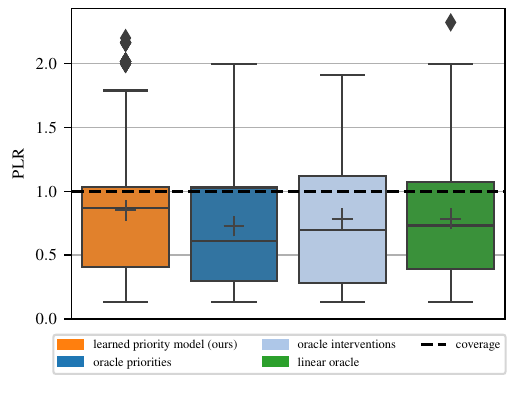}
    \caption{
        Performance results in the secondary target search task: Comparison of our method to oracle methods, displaying the episode-wise path length ratio (PLR) to the coverage baseline as boxplots.
    }
    \label{fig:boxplot_second}
    \vspace{-0.5em}
\end{figure}

\subsubsection{Qualitative Results}
\Cref{fig:exp_secondary_qual} presents an example scenario of the secondary target search task, comparing the coverage planner with our planner guided by learned priorities.
The target object is in the bathroom on the right side, and the robot starts in the bottom branch of the living room.
The living room connects to a large bedroom in the center and 3 kitchen-bathroom pairs at the top of the map.
The coverage robot incurs much performance loss when exploring the bedroom, while the learned semantic priorities favor continuing in the living room.
Both remaining paths in the upper part of the map are very similar, as the semantic features cannot strongly favor one direction over the other; all small rooms are semantically promising.
This example scenario indicates that the advantage of semantic over coverage exploration
is less pronounced in this scenario setup, as only the bedroom is a clearly semantically irrelevant area, while the remaining rooms are all prioritized. %

\subsubsection{Performance Results}

Quantitative performance results in the secondary task setup are presented in \cref{fig:boxplot_second}, analogous to \cref{sec:exp_performance}.
While our method outperforms the coverage planner (PLR $< 1$) in most episodes, the mean PLR of 0.853 is closer to 1 than in the primary task setup. 
This indicates more similar behavior of our method to the coverage planner, possibly as semantic priorities are less informative for target search.
This is also supported by the PLR boxplots of the oracle methods, showing that more episodes perform similar to coverage than in the primary setup.
Moreover, this task setup features a larger median gap between our approach and the oracle methods.
This shows that the difficulty of this task setup is exacerbated when using potentially noisy learned semantic priorities, giving more influence to the coverage gains in the tour cost function (\cref{eq:planner_priority}).
However, while some scenarios do not provide much room for improvement over coverage, the results show that our approach substantially improves target search efficiency in many other scenarios.
\looseness=-1

\section{Conclusion}

In this paper, we presented a novel approach to target search in unknown environments, combining semantic priorities learned from expert guidance with a global exploration planner.
We trained the semantic priority model weighting exploration frontiers based on semantic features, such that a derived expert model matches a dataset of expert interventions.
The combinatorial exploration planner prioritizes frontiers based on semantic priority and expected coverage gain, ensuring robust exploration independent of the learned model.
The results show that the exploration planner guided by the learned priority model exhibits efficient target search behavior and outperforms a purely coverage-driven planner variant across different scenarios and simulated expert datasets.
Future work will consider more realistic environments with complex semantic relationships and learning from real human data. \looseness=-1

\bibliographystyle{IEEEtran}
\bibliography{IEEEabrv,references}

\end{document}